\def\year{2021}\relax
\documentclass[letterpaper]{article} 
\usepackage{aaai21}  
\usepackage{times}  
\usepackage{helvet} 
\usepackage{courier}  
\usepackage[hyphens]{url}  
\usepackage{graphicx} 
\usepackage{natbib}  
\usepackage{caption} 
\usepackage{graphicx}
\usepackage{comment}
\usepackage{amsmath,amssymb} 

\usepackage{dsfont}
\usepackage{algorithm}
\usepackage{algorithmic}
\usepackage{makecell}
\usepackage{bm}
\usepackage{multirow}
\usepackage{soul}
\usepackage[caption=false]{subfig}
\usepackage[switch]{lineno}

\newcommand{\bd}[1]{\textbf{#1}}
\newcommand{\app}{\raise.17ex\hbox{$\scriptstyle\sim$}}

\newcolumntype{x}[1]{>{\centering\arraybackslash}p{#1pt}}

\newlength\savewidth
\newcommand{\tablestyle}[2]{\setlength{\tabcolsep}{#1}\renewcommand{\arraystretch}{#2}\centering\footnotesize}
\makeatletter\renewcommand\paragraph{\@startsection{paragraph}{4}{\z@}
  {.5em \@plus1ex \@minus.2ex}{-.5em}{\normalfont\normalsize\bfseries}}\makeatother

\title{SIMPLE: SIngle-network with Mimicking and Point Learning for  \\ Bottom-up Human Pose Estimation}
\author{Jiabin Zhang,\textsuperscript{\rm 1}\thanks{contributed equally}
        Zheng Zhu,\textsuperscript{\rm 2}\footnotemark[1]\thanks{corresponding author}
        Jiwen Lu,\textsuperscript{\rm 2}
        Junjie Huang,\textsuperscript{\rm 3}
        Guan Huang,\textsuperscript{\rm 3}
        Jie Zhou\textsuperscript{\rm 2}\\
}

\affiliations{
    \textsuperscript{\rm 1}Institute of Automation, Chinese Academy of Sciences, Beijing, China\\
    \textsuperscript{\rm 2}Tsinghua University, Beijing, China\\
    \textsuperscript{\rm 3}XForwardAI Technology Co.,Ltd, Beijing, China\\
    
    zhangjiabin2016@ia.ac.cn, \{zhengzhu, junjie.huang\}@ieee.org, \{lujiwen, jzhou\}@tsinghua.edu.cn, guan.huang@xforwardai.com

}

\begin{document}

\maketitle

\begin{abstract}
The practical application requests both accuracy and efficiency on multi-person pose estimation algorithms. But the high accuracy and fast inference speed are dominated by top-down methods and bottom-up methods respectively. To make a better trade-off between accuracy and efficiency, we propose a novel multi-person pose estimation framework, \textbf{SI}ngle-network with \textbf{M}imicking and \textbf{P}oint \textbf{L}earning for Bottom-up Human Pose \textbf{E}stimation (SIMPLE). Specifically, in the training process, we enable SIMPLE to mimic the pose knowledge from the high-performance top-down pipeline, which significantly promotes SIMPLE's accuracy while maintaining its high efficiency during inference. Besides, SIMPLE formulates human detection and pose estimation as a unified point learning framework to complement each other in single-network.  This is quite different from previous works where the two tasks may interfere with each other. To the best of our knowledge, both mimicking strategy between different method types and unified point learning are firstly proposed in pose estimation. In experiments, our approach achieves the new state-of-the-art performance among bottom-up methods on the COCO, MPII and PoseTrack datasets. Compared with the top-down approaches, SIMPLE has comparable accuracy and faster inference speed.
\end{abstract}

\section{Introduction}

Human pose estimation in images \cite{CPM,Hourglass} is of importance for visual understanding tasks \cite{app1,app2,app3}. Research community has witnessed a significant advance from single person to multi-person pose estimation, which can be generally categorized into top-down \cite{MSRAPose,hrnetv1}
and bottom-up \cite{DeepCut,OpenPose} approaches. Top-down methods achieve multi-person pose estimation by the two-stages process, including obtaining person bounding boxes by a person detector and predicting keypoint locations separately within these boxes. This time-consuming pipeline makes it much slower than bottom-up methods. Bottom-up methods firstly detect body joints without information of the number and locations of people. Then detected joints are grouped to form individual poses for person instances. However, the pose estimation of multi-people with different scales is difficult, so there is a noticeable performance gap between bottom-up and top-down methods.

\begin{table}[t]
\centering
\small
\begin{tabular}{ l| c | c }
Method & AP on COCO  & mAP on MPII \\ 
\Xhline{2\arrayrulewidth}
OpenPose & 61.8 & 75.6\\ 
AE & 65.5 & 79.6 \\  
PersonLab & 68.7 & - \\
MultiPoseNet & 69.6 & - \\
PPN  & - & 77.0\\
HigherHRNet  & 70.5 & - \\
Ours  & \textbf{71.1} & \textbf{85.1} \\%
\hline
\end{tabular}
\caption{The multi-person pose estimation performance for existing bottom-up methods and SIMPLE. Our approach obtains leading performance among bottom-up methods, with AP of 71.1 and mAP of 85.1 on COCO \texttt{test-dev} and MPII Human Pose \texttt{testing} sets respectively.}
\label{table:1}
\end{table}

\begin{figure*}[t]
\centering
\includegraphics[scale=0.20]{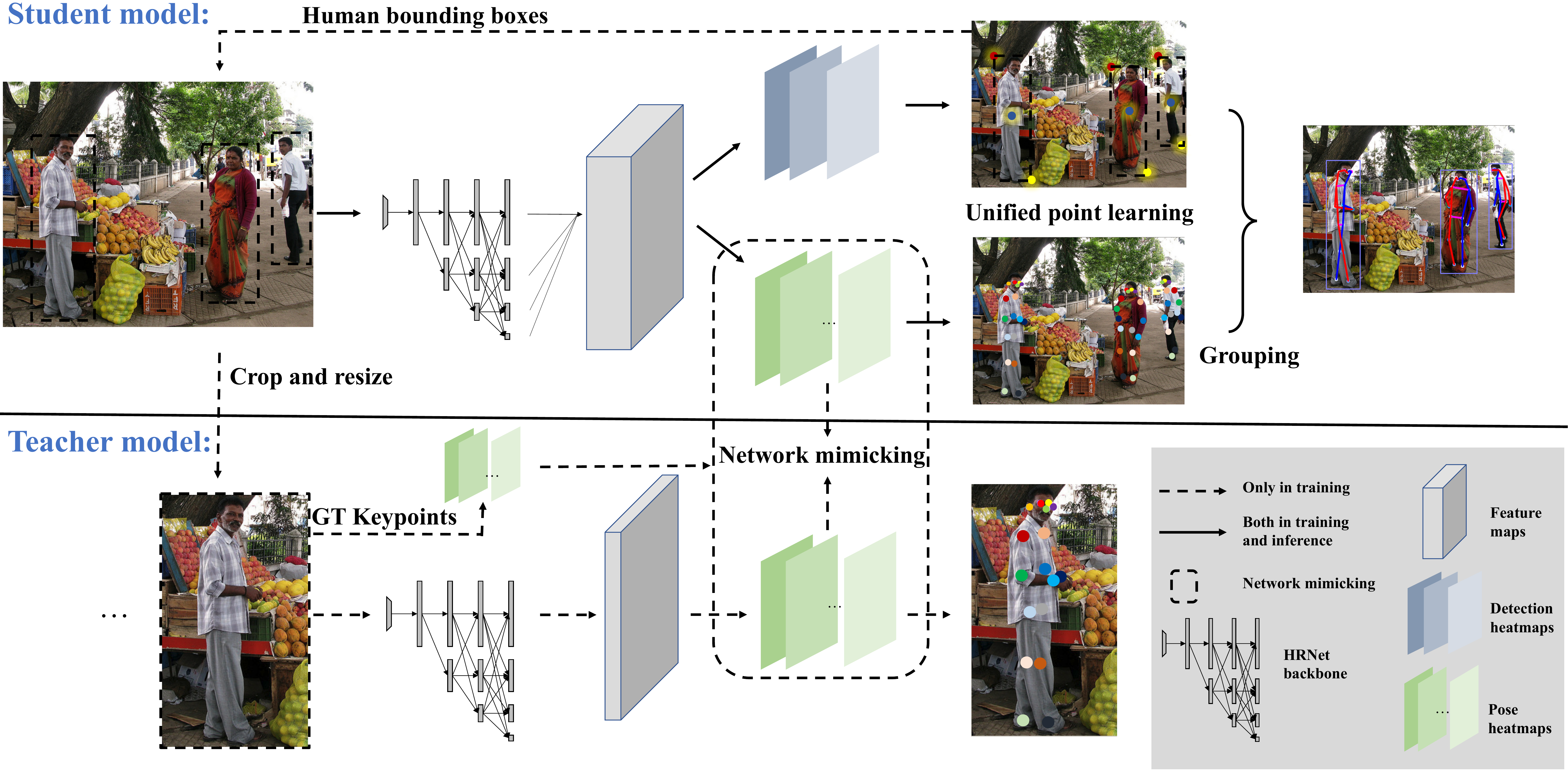}
\caption{The pipeline of SIMPLE. The upper half is the $student$ model, $i.e.$, our proposed bottom-up human pose estimation network, which has two point learning branches for human detection and pose estimation, respectively. The following module performs the keypoints grouping. The lower half is the $teacher$ model, a well-trained top-down pose estimation network. It performs single person pose estimation on the detected persons by the $student$ model and provides pose knowledge to improve the performance of $student$.}
\label{fig:net}
\end{figure*}

In this paper, we propose SIMPLE framework to address the problem mentioned above, i.e., there is not a satisfactory trade-off between accuracy and speed among the previous pose estimation methods. For narrowing the performance gap between bottom-up and top-down methods, SIMPLE forces the estimated pose heatmaps (student) to mimic the output heatmaps of the high-performance top-down approach (teacher) in the training process. Specifically, an extra human detection branch extracts the heatmap patches of each person in the student model, which is semantically aligned with the heatmaps of the teacher model. Then a convolutional adapter is utilized to align the high-level feature and the mimicking loss is optimized to perform knowledge transfer. For unifying human detection and pose estimation to complement each other, SIMPLE treats these two tasks both as point learning problems. In particular, it consists of a shared backbone and two parallel point learning-based head branches. Moreover, benefit from human detection results, SIMPLE has a CNN-based grouping module that can intuitively and graciously solve the grouping problem.

Benefit from the network mimicking strategy, SIMPLE is supervised by both ground-truth and high-level knowledge from the teacher model. Therefore its pose estimation performance is significantly promoted. Meanwhile, the fast inference speed is maintained. Point learning-based human detector and pose estimation not only promote each other but also help SIMPLE obtain an intuitive and gracious grouping algorithm. In experiments, SIMPLE achieves the new state-of-the-art performance among bottom-up methods on the COCO, MPII and PoseTrack dataset. Compared with the top-down approaches, SIMPLE has comparable accuracy and faster speed. As shown in Table \ref{table:1}, our approach outperforms previous bottom-up methods.

In conclusion, the main contributions of this paper can be described as follows:

1. SIMPLE proposes a novel network mimicking strategy to narrow the performance gap between bottom-up and top-down methods, which is the first work to adopt network mimicking between different pose estimation pipelines.

2. SIMPLE pioneers the unified point learning framework for both human detection and pose estimation, which makes the two tasks promote each other. Moreover, the detection results help SIMPLE achieve the network mimicking and solve the grouping problem intuitively and graciously.

3. SIMPLE achieves the new state-of-the-art performance among bottom-up methods on the COCO, MPII and PoseTrack dataset, which also has comparable accuracy with the top-down approaches but faster inference speed.

\section{Related Works}

\label{re-pose}

\paragraph{Multi-person Pose Estimation in Image}
Most top-down approaches \cite{GooglePose,CPN,MSRAPose,MSPN,hrnetv1,hrnetpami,udp,cvprun,rsn} achieve pose estimation by two-stages: firstly, detect and crop-resize persons from the original image, then perform single person pose estimation on fixed scale person patches. As the repetitive single person pose estimations of fixed size are performed for all the people in the image, most state-of-the-art performances on multi-person human pose estimation benchmarks are achieved by top-down methods. But this time-consuming pipeline makes it much slower than bottom-up methods. In contrast, bottom-up methods \cite{DeepCut,DeeperCut,OpenPose,ae,prn,higherhrnet} start by detecting identity-free joints for all persons in an input image through predicting heatmaps of different classes keypoints, and then group them into person instances. As the pose estimation only needs to be performed once, bottom-up methods can obtain faster inference speed regardless of the person numbers in an image. However, there is an obvious performance gap between bottom-up and top-down methods.

\paragraph{Point Learning for Visual Recognition}
Pose estimation is a typical application of point learning, which is performed only based on keypoint prediction. Object detection is another field where point learning is widely applied. Existing object detection methods can be categorized into two main types of pipelines: anchor-based and anchor-free approaches. Point learning-based object detection is a branch of anchor-free methods. CornerNet \cite{cornernet}, CenterNet \cite{objects, centernet} and CentripetalNet \cite{cpvrp1} are representative point learning detectors. Besides, point learning is recently developed in single \cite{zhang2020ocean} and multiple \cite{pointmot} object tracking.

\paragraph{Network Mimicking}
The principle of network mimicking is knowledge distillation. Concentrating on information transfer between different neural networks, knowledge distillation has been successfully exploited in many computer vision tasks such as classification \cite{hintondis, fitnets,mutual}, object detection \cite{disdet1eff,disdet3quan,dcnv2} and semantic segmentation \cite{affseg, disseg2}. While the previous knowledge distillation for pose estimation \cite{fastpose,dispose1} are implemented between the different configurations of the same pipeline, we propose a new knowledge distillation strategy to utilize the built-in advantage of the top-down method to improve the performance of bottom-up methods fundamentally.

\section{SIMPLE Framework}
\label{pipeline}

As shown in Fig. \ref{fig:net}, SIMPLE consists of a single network with point learning for bottom-up human pose estimation (described in Sec. \ref{sec_p}) and an assistant mimicking network for improving the pose estimation performance (described in Sec. \ref{sec_m}). As shown in the upper half of Fig. \ref{fig:net}, the pose estimator with point learning plays the $student$ model in the network mimicking. The input image is firstly fed into the feature extractor (backbone network) which is followed by two parallel branches including a human detection branch and pose estimation branch. The unified point learning pipelines complement each other due to their similar learning target. Additionally, the result of human detection can help SIMPLE assign the detected pose keypoints to achieve pose grouping.

\begin{figure*}[ht]
\centering
\includegraphics[scale=0.16]{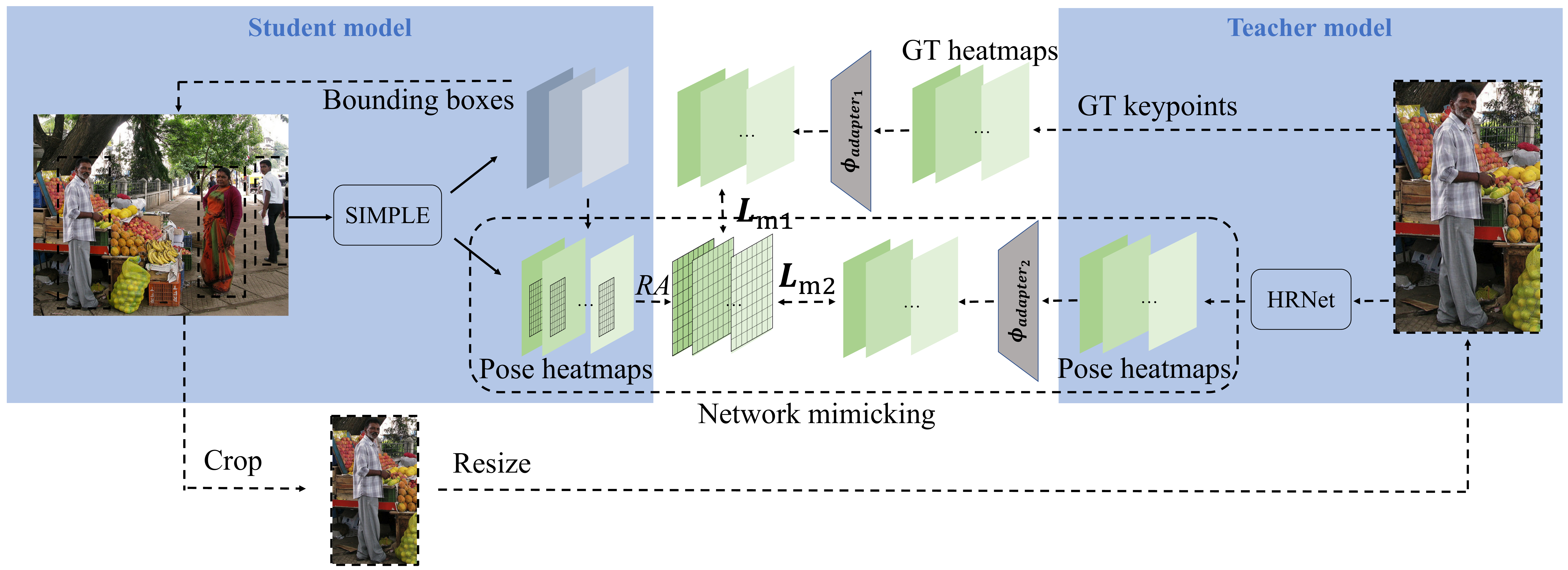}
\caption{The pipeline of network mimicking strategy. This figure is a part of Fig.  \ref{fig:net}, in which the content about network mimicking is reserved and drawn in more detail. The left blue area is our proposed $student$ model. Its detection branch provides bounding boxes to crop and resize the single person images which are fed into the top-down $teacher$ model in the right blue area. On the other hand, these bounding boxes are utilized to extract $student$ pose heatmaps of the detected person by $ROIAlign$ (RA). So mimicking performs between extracted $student$ pose heatmaps and the $teacher$ model's pose knowledge transferred by two convolutional adapters.}
\label{fig:mimic}
\end{figure*}

In the network mimicking strategy of SIMPLE, our proposed point learning network is the $student$ model. A high-performance top-down pose estimation method is adopted as the $teacher$ model which is drawn in the lower half of Fig. \ref{fig:net}. Network mimicking starts at the intermediate training process of the $student$ model. The bounding boxes outputted by the human detection branch of the $student$ model are utilized to crop and resize the sub-images of a single person. Then these single person images are fed into the top-down pipeline to perform single person pose estimation. Finally, network mimicking is performed between SIMPLE ($student$) and the top-down method ($teacher$) to improve the quality of pose heatmaps of SIMPLE, which is illustrated in Fig. \ref{fig:mimic}.

To keep the consistency of features between the $teacher$ and $student$ model, we use HRNet \cite{hrnetv1} as the backbones of both models. Inspired by \cite{hrnetv1}, the $teacher$ model only uses the feature maps of the highest-resolution branch in the 4th stage to predict the keypoints. For the proposed $student$ model which needs to predict the keypoints for all persons with different scales, we use all the outputs of four branches in the 4th stage. Specifically, up-sampling is performed to transform the resolution of other branches to the highest-resolution to obtain the feature maps $F_{shared} \in \mathcal{R}^{\frac{W}{4} \times \frac{H}{4} \times 15C}$. $C$ is the width of feature maps outputted by the 1st stage of HRNet.

\section{Point Learning for Human Detection and Pose Estimation}
\label{sec_p}
Different from the pre-processing stage in top-down methods, the human detector of SIMPLE is performed parallel with pose estimation. As shown in Fig. \ref{fig:net}, SIMPLE formulates the human detection and pose estimation as a unified point learning problem whose learning target is a set of points ${\mathcal P}_{SIMPLE}$:
\begin{scriptsize}
\begin{equation}
{\mathcal P}_{SIMPLE} = {\mathcal P}_{det}\bigcup{\mathcal P}_{pose}
\end{equation}
\end{scriptsize}

where ${\mathcal P}_{det}$ and ${\mathcal P}_{pose}$ are the two sets of human detector and pose estimator respectively.

\subsection{Point Learning for Human Detection}
As a subset of ${\mathcal P}_{SIMPLE}$, ${\mathcal P}_{det}$ is a union of three point sets:
\begin{scriptsize}
\begin{equation}
{\mathcal P}_{det} = \{{\mathcal P}_{top-left}, {\mathcal P}_{bottom-right}, {\mathcal P}_{center} \}
\end{equation}
\end{scriptsize}
where ${\mathcal P}_{top-left}$, ${\mathcal P}_{bottom-right}$ and ${\mathcal P}_{center}$ are the sets of top-left, bottom-up and center point of human bounding boxes respectively.  Taking ${\mathcal P}_{top-left}$ for example:
\begin{scriptsize}
\begin{equation}
{\mathcal P}_{top-left} = \{{P}_{top-left}^{n} \mid 1\leq n \leq N,  n\in \mathbb{Z} \}
\end{equation}
\end{scriptsize}
where $N$ is the number of persons in the image.

As the Fig. \ref{fig:net} shows, SIMPLE generates detection heatmaps $H_{center} \in \mathcal{R}^{\frac{W}{4} \times \frac{H}{4}}$, $H_{top-left} \in \mathcal{R}^{\frac{W}{4} \times \frac{H}{4}}$ and $H_{bottom-right} \in \mathcal{R}^{\frac{W}{4} \times \frac{H}{4}}$ by setting center pooling $Pooling_{cen}$ and cascade corner pooling $Pooling_{cor}$ on the feature maps $F_{shared} \in \mathcal{R}^{\frac{W}{4} \times \frac{H}{4} \times 15C}$. The center pooling is used to find the maximum value in its both horizontal and vertical directions and add them together. Then $H_{center} \in \mathcal{R}^{\frac{W}{4} \times \frac{H}{4}}$ is generated by a 3$\times$3 convolution layer $W_{center}$ following center pooling. Meanwhile $H_{top-left} \in \mathcal{R}^{\frac{W}{4} \times \frac{H}{4}}$ and $H_{bottom-right} \in \mathcal{R}^{\frac{W}{4} \times \frac{H}{4}}$ can also be generated by the same pipeline with cascade corner pooling instead of center pooling. The center pooling, cascade corner pooling and the training loss $\bm{L_{det}}$ of detection branch are all defined in the CenterNet. We suggest to refer to \cite{centernet} for details. The three sets of points can be obtained on its corresponding heatmap as:
\begin{scriptsize}
\begin{equation}
{\mathcal P}_{center} = \{{P}_{center}^{n} \mid H_{center}({P}_{center}^{n}) \geq {\phi}_{det} \}
\end{equation}
\end{scriptsize}
where ${\phi}_{det}$ is the threshold value for human detection prediction. Here we take ${\mathcal P}_{center}$ as an example. The points in ${\mathcal P}_{top-left}, {\mathcal P}_{bottom-right}, {\mathcal P}_{center}$ are drawn on the image with different colors in Fig. \ref{fig:net} for intuitive understanding.

Grouping top-$N$ center, top-left and bottom-right points to human bounding boxes follows two steps: (1) obtaining $N^2$ proposal boxes by combining top-$N$ top-left and bottom-right points; (2) For each proposal box, if there is a center point in its central region, it will be preserved. Otherwise, it will be removed. The confidence score of the bounding box is defined as the average scores of the three points.

\subsection{Point Learning for Pose Estimation}
Similar with human detector, the learning targets of pose estimator is also a union of of $K$ point sets:
\begin{scriptsize}
\begin{equation}
{\mathcal P}_{pose} = \{{\mathcal P}_{k} \mid 1\leq k \leq K,  k\in \mathbb{Z} \}
\end{equation}
\end{scriptsize}
where $K$ is the category number of human keypoints, $\mathbb{Z}$ represents the set of integer and ${\mathcal P}_{k}$ is the set of the $k$-th category human keypoints:
\begin{scriptsize}
\begin{equation}
{\mathcal P}_{k} = \{{P}_{k}^{n} \mid 1\leq n \leq N,  n\in \mathbb{Z} \}
\end{equation}
\end{scriptsize}
where $N$ is the number of the $k$-th category human keypoints in the image.

Consistent with other bottom-up approaches, the pose estimation branch of SIMPLE is designed to output the keypoint heatmaps $H_{pose} \in \mathcal{R}^{\frac{W}{4} \times \frac{H}{4} \times k}$ from the features maps $F_{shared} \in \mathcal{R}^{\frac{W}{4} \times \frac{H}{4} \times 15C}$ by fully-convolutional layers $W_{pose}$. Then we can obtain all $K$ subsets of ${\mathcal P}_{pose}$ by:
\begin{scriptsize}
\begin{equation}
{\mathcal P}_{k} = \{{P}_{k}^{n} \mid H_{pose}({P}_{k}^{n}) \geq {\phi}_{pose} \}
\label{getp}
\end{equation}
\end{scriptsize}
where ${\phi}_{pose}$ is the threshold value for human keypoints prediction. The points in every ${\mathcal P}_{k}$ are drawn on the image with different colors in Fig. \ref{fig:net} for intuitive understanding.

In training process, the Mean-Squared Error (MSE) function is adopted as training loss $L_{pose}$ of pose estimation:
\begin{scriptsize}
\begin{equation}
\bm{L_\text{pose}} =\frac{1}{K}\sum_{k=1}^{K}
\|{{\hat{H}}_{pose\ k}}- {H_{pose}}_{k}\|_2^2
\end{equation}
\end{scriptsize}
where ${{\hat{H}}_{pose\ k}}$ and ${H_{pose}}_{k}$ means the ground-truth and predicted heatmap of $k$-th joint respectively.

Benefit from bounding boxes provided by human detection branch, SIMPLE can achieve keypoint grouping by utilizing a convolutional module.
Specifically, this module takes a sub-region of global pose heatmap extracted by someone's bounding box as input. On the sub-heatmap, there may be some unexpected keypoint activations not belonged to this person. The convolutional module needs to suppress these activations and $\textbf{reserve (not refine)}$ the activations belonged to this person. So the training supervision is not ground truth keypoints but their corresponding predicted keypoints for this person.
Different from the Pose Residual Network (PRN) in MultiPoseNet \cite{prn} which uses a residual multilayer perception (MLP) to perform the same task, SIMPLE makes better use of spatial structures of human pose.


\subsection{Unified Point Learning}
As pose estimation and human detection are both formulated as a point learning strategy, SIMPLE achieves the two tasks by parallel branches. As shown in Fig. \ref{fig:net}, all the heatmaps of pose estimation and human detection are generated from the feature maps extracted by the shared backbone:
\begin{scriptsize}
\begin{equation}
\left\{
\begin{split}
&{H_{pose}}_{k} = W_{pose}(F_{shared}) \\
&{{H_{center}} = W_{center}({Pooling}_{cen}(F_{shared}))} \\
&{{H_{top-left}} = W_{top-left}({Pooling}_{cor}(F_{shared}))} \\
&{{H_{bottom-right}} = W_{bottom-right}({Pooling}_{cor}(F_{shared}))} \\
\end{split}
\right.
\end{equation}
\end{scriptsize}

Based on the above unified point learning strategy, the whole SIMPLE can be trained end-to-end in the supervision by ground-truth of human keypoints and bounding boxes:
\begin{scriptsize}
\begin{equation}
\label{train}
\textit{W}_{SIMPLE} = \mathop{\arg\min}_{W_{S}}(\bm{L_\text{pose}} + \alpha  \bm{L_\text{det}})
\end{equation}
\end{scriptsize}
where ${W}_{SIMPLE}$ represents the model weight of SIMPLE.

The reason why this human detection pipeline is adopted in SIMPLE is that the bounding boxes of humans are predicted only relying on point prediction. It means human detection and pose estimation's optimization objectives are consistent. On the other hand, the activation of joints on pose estimation heatmaps can make the shared feature maps useful for center pooling and cascade corner pooling. So the pose estimator can help human detector improve accuracy.  It is noting that in MultiPoseNet, the anchor-based detector and point learning-based pose estimation may interfere with each other. The experiment in Sec.\ref{ablation} also proves that unified point learning in SIMPLE is more superior.

\section{Network Mimicking}
\label{sec_m}

\subsection{Motivation}
\label{mo}
For the multi-person pose estimation, state-of-the-art performance is always achieved by top-down methods. The bottom-up methods are generally lagging behind the top-down methods in terms of accuracy on public datasets. The better performance of top-down methods comes from the repeated computation for each fixed scale person's pose estimation in the second stage, which can help the network learn the feature of human keypoint accurately without scale variation. Pose estimation network of bottom-up methods needs to predict the locations of joints of all people with various scales in once inference, which brings difficulty to extract high-quality features for keypoints location. To tackle this dilemma, network mimicking is adopted in SIMPLE, which forces the bottom-up heatmaps to mimic the knowledge of the high-performance top-down approach.

\subsection{Pose Knowledge}
\label{pk}
The key to performing network mimicking is designing the pose knowledge from the $teacher$ model. Firstly, considering that ground-truth heatmaps have accurate keypoint annotations in top-down framework, we force the single-person areas of the $student$'s heatmaps to mimic the ground-truth of the $teacher$. This produces marginal effect as shown in Sec.\ref{ablation}. Inspired by the well-explored knowledge strategy researcher, the soft output as well as ground-truth of the $teacher$ model are adopted.

For mimicking the soft output of the $teacher$ model, two choices are explored: feature maps and heatmaps. We argue that heatmaps are better because 1) Feature maps of the $teacher$ model only serve for pose estimation, while the $student$ model serves both detection and pose estimation. 2) Heatmaps include more explicit pose information which can help the $student$ model learn them more straightly. The experiment results in Table \ref{tab:ablation_nm} validate this choice.

\subsection{Transfer Module}
As shown in Fig. \ref{fig:mimic}, the transfer module to perform network mimicking includes three components: a $ROIAlign$ \cite{mask} and two convolutional adapters. The resolution of $teacher$'s knowledge is fixed because its input is the resized sub-image of a single person, while the resolution of the corresponding area on $student$ is various. Therefore a $ROIAlign$ operation is adopted to extract the area of each person on the pose heatmaps of $student$ model. Utilizing the results of the human detection branch of SIMPLE, the $student$ heatmaps to be distilled is generated, which have the same fixed size as the knowledge of $teacher$ model:
\begin{scriptsize}
\begin{equation}
\label{trans1}
{{H}^s_{n}} \in \mathcal{R}^{\frac{W}{4} \times \frac{H}{4} \times K} = ROIAlign_ {{bbox}_n}({H}^s)
\end{equation}
\end{scriptsize}
where ${H}^s$ is the pose heatmaps of SIMPLE, ${bbox}_n$ is the bounding box of $n$-th person.

Then the mimicked $teacher$'s heatmaps is calculated by:
\begin{scriptsize}
\begin{equation}
\label{trans2}
{H'}_{n}^{t_{gt}} \in \mathcal{R}^{\frac{W}{4} \times \frac{H}{4} \times K} = \bm\phi_{{adapter}_1}({H_{n}^{t_{gt}}})
\end{equation}
\begin{equation}
\label{trans3}
{H'}_{n}^{t_p} \in \mathcal{R}^{\frac{W}{4} \times \frac{H}{4} \times K} = \bm\phi_{{adapter}_2}({{H}}_{n}^{t_p})
\end{equation}
\end{scriptsize}
where ${H_{n}^{t_{gt}}}$ and $H_{n}^{t_p}$ are the ground-truth heatmaps and predicted heatmaps of $teacher$ model on the sub-image of $n$-th person. And the sub-image is cropped and resized by utilizing the results of the human detection branch of SIMPLE. The two convolutional adapters $\phi_{{adapter}_1}$ and $\phi_{{adapter}_2}$ are added on heatmaps to transfer the latent knowledge better. The obtained ${H'}_{n}^{t_{gt}}$ and ${H'}_{n}^{t_p}$ comprise the pose knowledge from $teacher$ model.

\subsection{Mimicking Pipeline}
An overview of the whole mimicking procedure is depicted in Fig. \ref{fig:mimic}. At first, we need to prepare a trained high-performance top-down pose estimator as the $teacher$ model. Then the $student$ model SIMPLE is trained until the detection results can be utilized for network mimicking. At last, the network mimicking procedure starts: the bounding boxes predicted by the SIMPLE are utilized to crop and resize the training image to get a single person's images. And these single person's images are fed into the $teacher$ model to obtain the pose knowledge, which includes predicted and ground-truth heatmaps.  The way to assign ground-truth (GT) keypoints to a single person's images is similar to that assigning GT keypoints to human proposals in Mask R-CNN. So $student$ model is forced to mimic these heatmaps,  and only the parameters weight of the $student$ model is updated during mimicking. This procedure is also summarised in the Algorithm \ref{alg:trainingprocess}.

\begin{algorithm}[h]
\label{mimic_p}
	\caption{Mimicking pipeline of SIMPLE}
	\begin{algorithmic}
		\REQUIRE ~ Already trained $teacher$ model $T$, $student$ model $S$ initialized by ImageNet pretraining\\
		\ENSURE ~ Trained $student$ model $S$ \\
		\STATE { \textbf{STAGE 1}: Training $student$ model SIMPLE $S$: }\\
		
		$\qquad$ {$\textit{W}_{S} = \mathop{\arg\min}_{W_{S}}(\bm{L_\text{pose}} + \alpha \bm{L_\text{det}})$}

		\STATE { \textbf{STAGE 2}: Start network mimicking:}\\
		
		$\qquad$ {$\textit{W}_{S} = \mathop{\arg\min}_{W_{S}}( \bm{L_\text{pose}} + \alpha \bm{L_\text{det}} + \beta \bm{L_\text{m}})$}
	\end{algorithmic}
	\label{alg:trainingprocess}
\end{algorithm}

As the pose knowledge from the $teacher$ model has two parts,  two Mean-Squared Error (MSE) functions are adopted as the mimicking loss:
\begin{scriptsize}
\begin{equation}
\bm{L_\text{m1}} = \frac{1}{N}\frac{1}{K}\sum_{n=1}^{N}  \sum_{k=1}^{K}
\|{{H}_{n}^s}_{k} - {{H'}_{n}^{t_{gt}}}_{k}\|_2^2
\end{equation}
\begin{equation}
\bm{L_\text{m2}} = \frac{1}{N}\frac{1}{K}\sum_{n=1}^{N}  \sum_{k=1}^{K}
\|{{H}_{n}^s}_{k} - {{H'}_{n}^{t_{p}}}_{k}\|_2^2
\end{equation}
\end{scriptsize}
where ${{H}_{n}^s}_{k}$, ${{H'}_{n}^{tgt}}_{k}$ and ${{H'}_{n}^{tp}}_{k}$ specify the extracted $student$ heatmaps and transferred $teacher$ heatmaps for the $k$-th joint of the $n$-th people, calculated by Equation \ref{trans1}, Equation \ref{trans2} and Equation \ref{trans3} respectively. The final mimicking loss $\bm{L_\text{m}} $ is the sum of $\bm{L_\text{m1}}$ and $\bm{L_\text{m2}}$.

\subsection{Discussions}
Here are some explanations for why the mimicking strategy can help SIMPLE improve its performance:
(1) The predicted heatmaps of a high-performance $teacher$ network encode the abstract knowledge learned from the entire training dataset in advance. (2) The predicted heatmaps have more useful information (like more complete joint labels) about difficult training cases, and can mitigate inconsistencies (even some errors) in manual annotation \cite{fastpose}. (3) The pose knowledge from scale-invariance top-down method can help the bottom-up method mitigate scale variation problem.

\section{Experiments}

\begin{figure*}[t]
\centering
\includegraphics[scale=0.35]{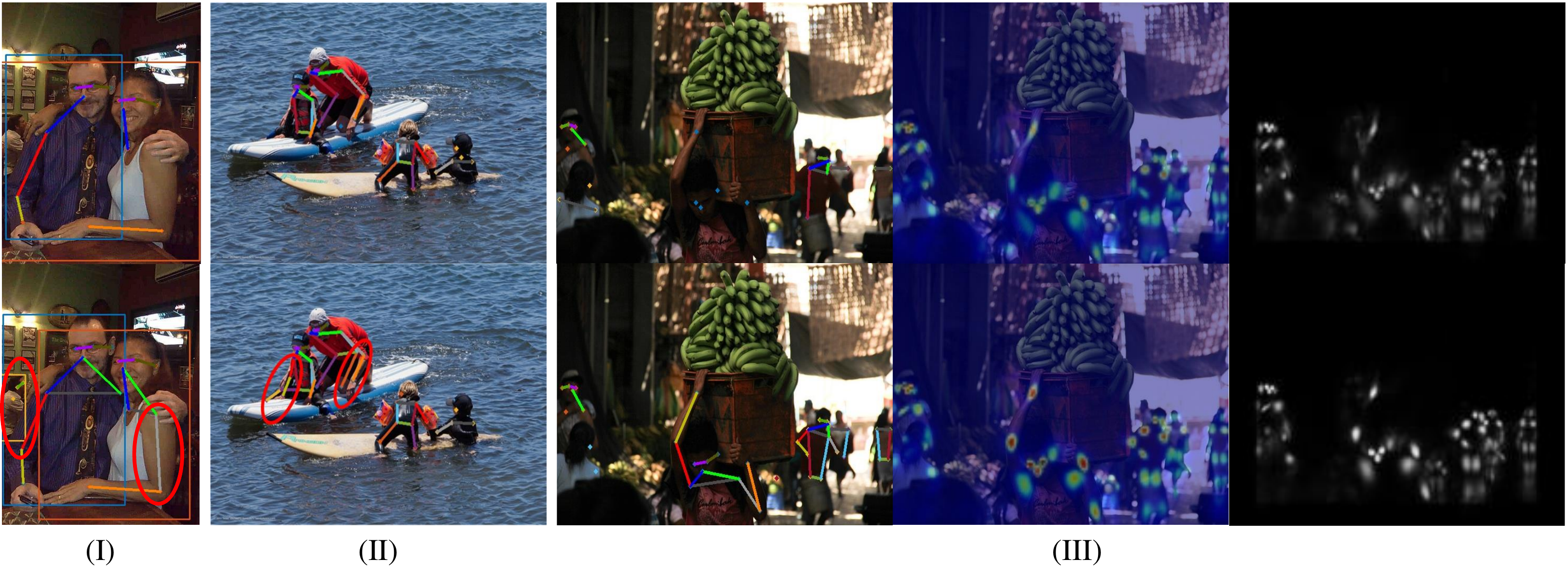}
\caption{Some pose estimation results on COCO2017 \texttt{val} dataset. (I) The upper and lower results are achieved by SIMPLE with regression-based and point learning-based human detections respectively, and improvements are marked by red circles. (II) The upper and lower results are achieved by SIMPLE-W32 without and with network mimicking respectively. (III) Columns 1$-$3 are pose estimation results, the combinations of the original image and colourful pose heatmaps, predicted pose heatmaps respectively. The upper and lower lines are performed by SIMPLE-W32 without and with network mimicking respectively.}
\label{fig:vis_all}
\end{figure*}

\subsection{Implemented Details}
The proposed SIMPLE is implemented in PyTorch~\cite{Pytorch}. Different specifications of HRNet are chosen as the backbone, including HRNet-W18 and HRNet-W32. They are all pre-trained on ImageNet~\cite{ImageNet}. The training process is performed on 8 GeForce 1080Ti (11GB) GPUs. The SIMPLE-W18 and SIMPLE-W32 are trained with a learning rate of 0.00025 and 0.0002 in the first 120 epochs while the batch size is set as 48 and 32, respectively. Then the learning rate drops 10 $\times$ for the 121-th to 200-th epochs. The network mimicking begins from 151-th epoch with the constant learning rate, while $\alpha$ and $\beta$ are set as 1. In inference, the size of the test image is set as 512 $\times$ 512. For the human detection branch, the $N$ is set as 32 when grouping top-N center, top-left and bottom-right points to human bounding boxes. Then we also select top-32 bounding boxes to finish pose keypoints grouping. 
For the multi-scale testing, the test image resolutions are scaled with 0.6, 1.0, 1.2, 1.5 and 1.8.

\subsection{Ablation Study}
\label{ablation}

\paragraph{Ablation for Point Learning Detection} In Table \ref{tab:ablation_ponit}, we evaluate the effectiveness of point learning detection in pose estimation framework by comparing: (a) SIMPLE without human detection, (b) SIMPLE with regression-based human-detection and (c) SIMPLE with point learning-based human detection. HRNet-W32 is utilized as the backbones of all three implementations. Additionally, due to the lack of human detect results, keypoint grouping of SIMPLE cannot be performed in (a). So the grouping method of AE \cite{ae} is adopted in (a). Network mimicking is not adopted in this ablation. In (b), the human detection is achieved as in MultiPoseNet. As reported in Table \ref{tab:ablation_ponit}, adopting regression-based human detection increases AP of 1.1 than that without human detection. Benefit from point learning human detection, our SIMPLE-W32 obtains AP of 67.1 (+2.6 AP than regression-based human detection) on COCO2017 \texttt{val}. This experiment proves the effectiveness of the unified point learning strategy. This can also be seen in Fig. \ref{fig:vis_all}.(I) that shows the visualizations of (b) and (c).

\begin{table}[t!]
    \centering
    \begin{tabular}{c|c|c|c}
    \hline
     & Detection & Point learning detection & AP  \\
    \Xhline{2\arrayrulewidth}
    (a) & & & 63.4 \\
    (b) & \checkmark & & 64.5 \\
    (c) & \checkmark & \checkmark & \textbf{67.1}\\
    \hline
    \end{tabular}
    \caption{Ablation study of point learning strategy. Trained on COCO2017 \texttt{train} and tested on COCO2017 \texttt{val}.}
    \label{tab:ablation_ponit}
\end{table}

\begin{table}[t!]
\centering
\begin{tabular}{ c|c|c }
\hline
 & Network mimicking strategy & AP  \\
\Xhline{2\arrayrulewidth}
(a) & Without network mimicking & 67.1 \\
(b) & Only with ground-truth heatmaps & 67.9  \\
(c) & Mimicking on soft feature maps  & 66.8 \\
(d) & Mimicking on soft heatmaps & 70.2 \\
(e) & SIMPLE's implementation & \textbf{70.8}\\
\hline
\end{tabular}
\caption{Ablation study of network mimicking strategy. Trained on COCO2017 \texttt{train} and tested on COCO2017 \texttt{val}.}
\label{tab:ablation_nm} 
\end{table}

\paragraph{Ablation for Network Mimicking} We evaluate the effectiveness of SIMPLE's mimicking strategy in Table \ref{tab:ablation_nm}. The $student$ model is SIMPLE-W32. (a) is without mimicking. As described in Sec. \ref{pk}, we firstly let SIMPLE's pose heatmaps mimic the ground-truth heatmaps of the top-down $teacher$ model. But the AP improvement of 0.8 is marginal as shown in Table \ref{tab:ablation_nm} (b). Then the soft output of the $teacher$ model is adopted, including feature maps (c) and predicted heatmaps (d). From Table \ref{tab:ablation_nm}, one can find that mimicking feature maps harms SIMPLE's accuracy while mimicking heatmaps can help (a) increase AP by 3.1. The final mimicking strategy (e) is a combination of (b) and (d), and the AP improvement of 3.7 proves the superiority of SIMPLE's strategy. 
Some visualization results are shown in Fig. \ref{fig:vis_all}.(II), which intuitively shows the improvement brought by mimicking. Additionally, heatmaps analysis is also illustrated in Fig. \ref{fig:vis_all}.(III). The visualization process of pose heatmaps is: firstly, we turn the pose heatmaps of 17 channels into a one-channel heatmap, where the pixel value at each location is the maximum among these of 17 channels. Then the one-channel pose heatmap is multiplied by 255. As shown in Fig. \ref{fig:vis_all}.(III), the predicted pose heatmaps with network mimicking are clearer and have higher activation value for the difficult joints than these without network mimicking.

\subsection{Comparison with State-of-the-art}

We compare SIMPLE with the state-of-the-art methods on the three datasets: COCO2017 \cite{coco}, MPII \cite{MPII} and PoseTrack \cite{PoseTrackBenchmark}. The compared bottom-up pose estimation methods include OpenPose \cite{OpenPose}, AE, PersonLab \cite{personlab}, MultiPoseNet, PifPaf \cite{PifPaf}, SPM  \cite{SPM}, HigherHRNet \cite{higherhrnet} and PPN \cite{posepartition}. The compared top-down pose estimation methods include Mask R-CNN \cite{mask}, CPN \cite{CPN}, RMPE\cite{rmpe}, SimpleBaseline \cite{MSRAPose} and HRNet-W32 \cite{hrnetv1}. The comparison pose tracking methods include JointFlow \cite{JointFlow}, PoseTrack \cite{PoseTrackBenchmark}, PoseFlow \cite{PoseFlow}, SimpleBaseline \cite{MSRAPose} and Det$\&$Track \cite{Det_and_Track}.

\begin{table}[t!]
\tablestyle{4pt}{1.0}
\centering
\begin{tabular}[t]{l|x{16}|x{16}x{12}x{12}x{12}x{12}x{24}} %
\hline
\multirow{2}{*}{Method}& \multirow{2}{*}{Type} &  \bd{AP} & AP & AP & AP & AP  & \bd{Speed} \\%
 &  & \bd{Total} & 50 & 75 & M& L &\bd{s/img} \\ %
\hline
 OpenPose ($r$) & BU & 61.8 & 84.9 & 67.5 & 57.1& 68.2  & 0.005 \\ 
 AE ($mr$) & BU & 65.5 & 86.8 & 72.3 & 60.6& 72.6  & 0.17   \\ 
 PersonLab ($m$) & BU & 68.7 & 89.0 & 75.4 & 64.1& 75.5  & -   \\ 
 MultiPoseNet ($mcer$) & BU & 69.6 & 86.3 & 76.6 & 65.0& 76.3  & 0.043   \\ 
 PifPaf & BU & 66.7 & - & - & 62.4& 72.9  & 0.4   \\ 
 SPM & BU & 66.9 & 88.5 & 72.9 & 62.6& 73.1  & 0.058   \\ 
 HigherHRNet-32  & BU &  66.4 & 87.5 & 72.8 & 61.2 & 74.2& - \\
 HigherHRNet-48 ($m$) & BU &  70.5 & 89.3 & 77.2 & 66.6 & 75.8 & - \\
 SIMPLE-W18 & BU & 66.1 & 85.9 & 73.6 & 63.6 & 74.6& 0.040   \\ %
 SIMPLE-W18 ($m$) & BU & 68.1 & 87.8 & 75.5 & 64.9& 75.7  & 0.040   \\ %
 SIMPLE-W32  & BU & 69.6 & 89.3 & 77.9 & 68.1& 77.8  & 0.072   \\ %
 SIMPLE-W32  ($m$) & BU & \textbf{71.1} & 90.2 & 79.4 & 69.1& 79.1  & 0.072 \\  \hline%
 Mask R-CNN & TD & 63.1 & 87.3 & 68.7 & 57.8& 71.4  & 0.2 \\
 CPN & TD & 73.0 & 91.7 & 80.9 & 69.5& 78.1  & -\\
 RMPE & TD & 68.8 & 87.5 & 75.9 & 64.6  & 75.1 & 2.5\\
 SimpleBaseline & TD & 73.7 & 91.9 & 81.1 & 70.3& 80.0  &  5  \\
 HRNet-W32 & TD & \textbf{74.9} & 92.5 & 82.8 & 71.3 & 80.9 & -\\
 \hline
\end{tabular}
\caption{Multi-person pose estimation performance on COCO2017 \texttt{test-dev} dataset. $m$, $c$, $e$ and $r$ are the usual data augmentations in the bottom-up pipeline. $m$ means multi scale testing. $c$ means multi-crop testing. $e$ means two different network are used for ensemble. $r$ means using a single person pose refinement network. }
\label{tab:coco}
\end{table}

\paragraph{COCO2017} Table \ref{tab:coco} reports the results of pose estimation on COCO2017 \texttt{test-dev} dataset. Our SIMPLE-W32 obtains AP of 71.1, which outperforms all the bottom-up methods. Using a larger backbone HRNet-W48 and larger input size of 640, HigherHRNet still lags behind SIMPLE-W32. Even without test data augmentations, SIMPLE-W32 can also achieve the almost 70 AP score, which is significantly higher than HigherHRNet with the same configuration. Benefit from our unified point learning and network mimicking strategy, SIMPLE still outperforms most bottom-up methods with data augmentations. Compared with top-down methods, SIMPLE achieves comparable AP, meanwhile much faster than them (the input size of SIMPLE is 512$\times$512). Comparing with SimpleBaseline, SIMPLE-W32 increases in speed by dozens of times while is only 2.6 AP behind it.

\paragraph{MPII}
Table \ref{tab:mpii} reports the results on MPII. Only bottom-up methods participate in the comparison because there are no bounding box annotations for top-down methods. On MPII Human Pose \texttt{testing} set, the mAP of the two SIMPLEs are 82.5 and 85.1. These performances beat all the previous bottom-up methods that reported the result on MPII dataset in their paper.

\begin{table}[t]
\tablestyle{4pt}{1.0}
\begin{tabular}{l|l|x{12}x{9}x{9}x{9}x{9}x{9}x{9}x{9}}
\hline
\multirow{2}{*}{Method}& \multirow{2}{*}{Type} &  \bd{mAP} & AP & AP & AP & AP & AP & AP & AP \\
 &  & \bd{Total} & Hea & Sho & Elb & Wri & Hip & Kne & Ank \\
\hline
 OpenPose ($r$) & BU & 75.6 & 91.2 & 87.6 & 77.7 & 66.8& 75.4 & 68.9 & 61.7    \\ 
 AE ($mr$)  & BU & 77.5 & 92.1 & 89.3 & 78.9 & 69.8  & 76.2 & 71.6 & 64.7\\ 
 PPN ($mcer$)  & BU & 80.4 & 92.2 & 89.7 & 82.1 & 74.4 & 78.6 & 76.4 & 69.3   \\ 
 SIMPLE-W18 & BU & 81.4 &94.9 & 91.5 & 83.7 & 74.3 & 81.9 & 73.6 & 69.8   \\ %
 SIMPLE-W18 ($m$) & BU & 82.5 & 96.8 & 92.8 & 85.0 & 75.5  & 83.2& 74.0 & 70.2  \\ %
 SIMPLE-W32  & BU & 83.3 & 97.6 & 93.2 & 85.6 & 76.8 & 83.9 & 74.6 & 71.0  \\ %
 SIMPLE-W32 ($m$) & BU & \textbf{85.1} & 98.4 & 94.6 & 87.2 & 78.2& 85.8 & 77.1 & 73.6   \\  %
 \hline
\end{tabular}
\caption{Multi-person pose estimation on MPII Human Pose \texttt{testing} dataset.} 
\label{tab:mpii}
\end{table}

\begin{table}[h]
\tablestyle{4pt}{1.0}
\begin{tabular}{l|l|x{18}x{20}x{20}x{20}x{20}}
\hline
\multirow{2}{*}{Method}& \multirow{2}{*}{Type} &  \bd{mAP} & \bd{MOTA} & Prec & Rec & \bd{Speed} \\
 &  &
 \bd{Total} & \bd{Total}& Total & Total & \bd{s/img} \\
 \hline
 JointFlow  & BU &  63.3 & 53.1  & 82.6 &  69.7   & 5  \\ 
 PoseTrack  & BU &  59.4 & 48.4 & - & - & -  \\
 SIMPLE-W18  & BU & 66.7 & 55.1 & 77.3 & 70.3 & 0.040 \\
 SIMPLE-W32  & BU & \textbf{69.5} & \textbf{55.7} & 79.1 & 74.6 & 0.072 \\ \hline
 PoseFlow & TD &  63.0 & 51.0 & 71.2 & 78.9 & 0.15    \\
 Sim.Bas.-50 & TD &  70.0 & 56.4 & 81.0 & 75.7 & 5 \\
 Sim.Bas.-152 & TD &  \textbf{73.9} & \textbf{57.6}& 79.4 & 79.9 & 5 \\
 Det$\&$Track & TD &  59.6 & 51.8 & - & - & 1.25 \\
 \hline
\end{tabular}
\caption{Multi-person pose estimation and tracking performance on PoseTrack \texttt{test} dataset.}
\label{tab:posetrack}
\end{table}

\paragraph{PoseTrack} PoseTrack is a large-scale benchmark for human pose estimation and tracking in videos. mAP and MOTA are two main metrics for pose estimation and tracking performance. Table \ref{tab:posetrack} reports the results on the \texttt{test} dataset. The pose estimation on each video frame is achieved by SIMPLE, and pose tracking strategy across frames follows SimpleBaseline, which can be referred to \cite{MSRAPose}. As reported in Table \ref{tab:posetrack}, our SIMPLE-W32 achieves the competitive result, a 69.5 mAP and a 55.7 MOTA. The results are only weaker than two SOTA top-down methods and achieves leading performance among the bottom-up methods. It's worth noting that the inference speed of SIMPLE-W32 is only 0.072 s/image, which is significant for the practicality of pose tracking.

\section{Conclusion}
In this paper, we propose SIMPLE, a novel multi-person pose estimation framework. SIMPLE formulates the human detection and pose estimation as a unified point learning problem, and can mimic knowledge between two different pipelines. In experiments, SIMPLE achieves better accuracy-speed trade-off comparing with SOTAs.

\bibliographystyle{aaai21}
\bibliography{egbib}

\begin{thebibliography}{50}
\providecommand{\natexlab}[1]{#1}
\providecommand{\url}[1]{\texttt{#1}}
\providecommand{\urlprefix}{URL }
\expandafter\ifx\csname urlstyle\endcsname\relax
  \providecommand{\doi}[1]{doi:\discretionary{}{}{}#1}\else
  \providecommand{\doi}{doi:\discretionary{}{}{}\begingroup
  \urlstyle{rm}\Url}\fi

\bibitem[{Andriluka et~al.(2018)Andriluka, Iqbal, Milan, Insafutdinov,
  Pishchulin, Gall, and Schiele}]{PoseTrackBenchmark}
Andriluka, M.; Iqbal, U.; Milan, A.; Insafutdinov, E.; Pishchulin, L.; Gall,
  J.; and Schiele, B. 2018.
\newblock Posetrack: A benchmark for human pose estimation and tracking.
\newblock In \emph{IEEE Conference on Computer Vision and Pattern Recognition},
  5167--5176.

\bibitem[{Andriluka et~al.(2014)Andriluka, Pishchulin, Gehler, and
  Schiele}]{MPII}
Andriluka, M.; Pishchulin, L.; Gehler, P.; and Schiele, B. 2014.
\newblock 2d human pose estimation: New benchmark and state of the art
  analysis.
\newblock In \emph{IEEE Conference on Computer Vision and Pattern Recognition},
  3686--3693.

\bibitem[{Artacho et~al.(2020)Artacho, Savakis, Savakis, and Savakis}]{cvprun}
Artacho, B.; Savakis, A.; Savakis, A.; and Savakis, A. 2020.
\newblock UniPose: Unified human pose estimation in single images and videos.
\newblock In \emph{IEEE Conference on Computer Vision and Pattern Recognition},
  7035--7044.

\bibitem[{Cai et~al.(2020)Cai, Wang, Luo, Yin, Du, Wang, Zhou, Zhou, Zhang, and
  Sun}]{rsn}
Cai, Y.; Wang, Z.; Luo, Z.; Yin, B.; Du, A.; Wang, H.; Zhou, X.; Zhou, E.;
  Zhang, X.; and Sun, J. 2020.
\newblock Learning delicate local representations for multi-Person pose
  estimation.
\newblock In \emph{European Conference on Computer Vision}.

\bibitem[{Cao et~al.(2019)Cao, Hidalgo, Simon, Wei, and Sheikh}]{OpenPose}
Cao, Z.; Hidalgo, G.; Simon, T.; Wei, S.-E.; and Sheikh, Y. 2019.
\newblock OpenPose: realtime multi-person 2D pose estimation using part
  affinity fields.
\newblock \emph{IEEE Transactions on Pattern Analysis and Machine Intelligence}
  .

\bibitem[{Chen et~al.(2018)Chen, Wang, Peng, Zhang, Yu, and Sun}]{CPN}
Chen, Y.; Wang, Z.; Peng, Y.; Zhang, Z.; Yu, G.; and Sun, J. 2018.
\newblock Cascaded pyramid network for multi-person pose estimation.
\newblock In \emph{IEEE Conference on Computer Vision and Pattern Recognition},
  7103--7112.

\bibitem[{Cheng et~al.(2020)Cheng, Xiao, Wang, Shi, Huang, and
  Zhang}]{higherhrnet}
Cheng, B.; Xiao, B.; Wang, J.; Shi, H.; Huang, T.~S.; and Zhang, L. 2020.
\newblock HigherHRNet: Scale-aware representation learning for bottom-up human
  pose estimation.
\newblock In \emph{IEEE Conference on Computer Vision and Pattern Recognition},
  5386--5395.

\bibitem[{Doering, Iqbal, and Gall(2018)}]{JointFlow}
Doering, A.; Iqbal, U.; and Gall, J. 2018.
\newblock Joint flow: Temporal flow fields for multi person tracking.
\newblock In \emph{British Machine Vision Conference}.

\bibitem[{Dong et~al.(2020)Dong, Li, Liao, Wang, Ren, and Qian}]{cpvrp1}
Dong, Z.; Li, G.; Liao, Y.; Wang, F.; Ren, P.; and Qian, C. 2020.
\newblock CentripetalNet: Pursuing high-quality keypoint pairs for object
  detection.
\newblock In \emph{IEEE Conference on Computer Vision and Pattern Recognition},
  10519--10528.

\bibitem[{Duan et~al.(2019)Duan, Bai, Xie, Qi, Huang, and Tian}]{centernet}
Duan, K.; Bai, S.; Xie, L.; Qi, H.; Huang, Q.; and Tian, Q. 2019.
\newblock Centernet: Keypoint triplets for object detection.
\newblock In \emph{IEEE International Conference on Computer Vision},
  6569--6578.

\bibitem[{Fang et~al.(2017)Fang, Xie, Tai, and Lu}]{rmpe}
Fang, H.-S.; Xie, S.; Tai, Y.-W.; and Lu, C. 2017.
\newblock Rmpe: Regional multi-person pose estimation.
\newblock In \emph{IEEE International Conference on Computer Vision},
  2334--2343.

\bibitem[{Girdhar et~al.(2018)Girdhar, Gkioxari, Torresani, Paluri, and
  Tran}]{Det_and_Track}
Girdhar, R.; Gkioxari, G.; Torresani, L.; Paluri, M.; and Tran, D. 2018.
\newblock Detect-and-track: efficient pose estimation in videos.
\newblock In \emph{IEEE Conference on Computer Vision and Pattern Recognition},
  350--359.

\bibitem[{Hattori et~al.(2018)Hattori, Lee, Boddeti, Beainy, Kitani, and
  Kanade}]{app1}
Hattori, H.; Lee, N.; Boddeti, V.~N.; Beainy, F.; Kitani, K.~M.; and Kanade, T.
  2018.
\newblock Synthesizing a scene-specific pedestrian detector and pose estimator
  for static video surveillance.
\newblock \emph{International Journal of Computer Vision} 126(9): 1027--1044.

\bibitem[{He et~al.(2017)He, Gkioxari, Doll{\'a}r, and Girshick}]{mask}
He, K.; Gkioxari, G.; Doll{\'a}r, P.; and Girshick, R. 2017.
\newblock Mask r-cnn.
\newblock In \emph{IEEE International Conference on Computer Vision},
  2961--2969.

\bibitem[{He et~al.(2019)He, Shen, Tian, Gong, Sun, and Yan}]{affseg}
He, T.; Shen, C.; Tian, Z.; Gong, D.; Sun, C.; and Yan, Y. 2019.
\newblock Knowledge adaptation for efficient semantic segmentation.
\newblock In \emph{IEEE Conference on Computer Vision and Pattern Recognition},
  578--587.

\bibitem[{Hinton, Vinyals, and Dean(2015)}]{hintondis}
Hinton, G.; Vinyals, O.; and Dean, J. 2015.
\newblock Distilling the knowledge in a neural network.
\newblock \emph{arXiv preprint arXiv:1503.02531} .

\bibitem[{Huang et~al.(2020)Huang, Zhu, Guo, and Huang}]{udp}
Huang, J.; Zhu, Z.; Guo, F.; and Huang, G. 2020.
\newblock The devil is in the details: Delving into unbiased data processing
  for human pose estimation.
\newblock In \emph{IEEE Conference on Computer Vision and Pattern Recognition},
  5700--5709.

\bibitem[{Insafutdinov et~al.(2016)Insafutdinov, Pishchulin, Andres, Andriluka,
  and Schiele}]{DeeperCut}
Insafutdinov, E.; Pishchulin, L.; Andres, B.; Andriluka, M.; and Schiele, B.
  2016.
\newblock Deepercut: A deeper, stronger, and faster multi-person pose
  estimation model.
\newblock In \emph{European Conference on Computer Vision}, 34--50.

\bibitem[{Kocabas, Karagoz, and Akbas(2018)}]{prn}
Kocabas, M.; Karagoz, S.; and Akbas, E. 2018.
\newblock Multiposenet: Fast multi-person pose estimation using pose residual
  network.
\newblock In \emph{European Conference on Computer Vision}, 417--433.

\bibitem[{Kreiss, Bertoni, and Alahi(2019)}]{PifPaf}
Kreiss, S.; Bertoni, L.; and Alahi, A. 2019.
\newblock Pifpaf: Composite fields for human pose estimation.
\newblock In \emph{IEEE Conference on Computer Vision and Pattern Recognition},
  11977--11986.

\bibitem[{Law and Deng(2018)}]{cornernet}
Law, H.; and Deng, J. 2018.
\newblock Cornernet: Detecting objects as paired keypoints.
\newblock In \emph{European Conference on Computer Vision}, 734--750.

\bibitem[{Li et~al.(2018)Li, Chen, Zhang, and Huang}]{app2}
Li, D.; Chen, X.; Zhang, Z.; and Huang, K. 2018.
\newblock Pose guided deep model for pedestrian attribute recognition in
  surveillance scenarios.
\newblock In \emph{IEEE International Conference on Multimedia and Expo}, 1--6.

\bibitem[{Li, Jin, and Yan(2017)}]{disdet1eff}
Li, Q.; Jin, S.; and Yan, J. 2017.
\newblock Mimicking very efficient network for object detection.
\newblock In \emph{IEEE Conference on Computer Vision and Pattern Recognition},
  6356--6364.

\bibitem[{Li et~al.(2019)Li, Wang, Yin, Peng, Du, Xiao, Yu, Lu, Wei, and
  Sun}]{MSPN}
Li, W.; Wang, Z.; Yin, B.; Peng, Q.; Du, Y.; Xiao, T.; Yu, G.; Lu, H.; Wei, Y.;
  and Sun, J. 2019.
\newblock Rethinking on multi-stage networks for human pose estimation.
\newblock \emph{arXiv preprint arXiv:1901.00148} .

\bibitem[{Lin et~al.(2014)Lin, Maire, Belongie, Hays, Perona, Ramanan,
  Doll{\'a}r, and Zitnick}]{coco}
Lin, T.-Y.; Maire, M.; Belongie, S.; Hays, J.; Perona, P.; Ramanan, D.;
  Doll{\'a}r, P.; and Zitnick, C.~L. 2014.
\newblock Microsoft COCO: Common objects in context.
\newblock In \emph{European Conference on Computer Vision}, 740--755.

\bibitem[{Liu et~al.(2019)Liu, Chen, Liu, Qin, Luo, and Wang}]{disseg2}
Liu, Y.; Chen, K.; Liu, C.; Qin, Z.; Luo, Z.; and Wang, J. 2019.
\newblock Structured knowledge distillation for semantic segmentation.
\newblock In \emph{IEEE Conference on Computer Vision and Pattern Recognition},
  2604--2613.

\bibitem[{Newell, Huang, and Deng(2017)}]{ae}
Newell, A.; Huang, Z.; and Deng, J. 2017.
\newblock Associative embedding: End-to-end learning for joint detection and
  grouping.
\newblock In \emph{Advances in Neural Information Processing Systems},
  2277--2287.

\bibitem[{Newell, Yang, and Deng(2016)}]{Hourglass}
Newell, A.; Yang, K.; and Deng, J. 2016.
\newblock Stacked hourglass networks for human pose estimation.
\newblock In \emph{European Conference on Computer Vision}, 483--499.

\bibitem[{Nie et~al.(2018)Nie, Feng, Xing, and Yan}]{posepartition}
Nie, X.; Feng, J.; Xing, J.; and Yan, S. 2018.
\newblock Pose partition networks for multi-person pose estimation.
\newblock In \emph{European Conference on Computer Vision}, 684--699.

\bibitem[{Nie et~al.(2019)Nie, Feng, Zhang, and Yan}]{SPM}
Nie, X.; Feng, J.; Zhang, J.; and Yan, S. 2019.
\newblock Single-stage multi-person pose machines.
\newblock In \emph{IEEE International Conference on Computer Vision},
  6951--6960.

\bibitem[{Papandreou et~al.(2018)Papandreou, Zhu, Chen, Gidaris, Tompson, and
  Murphy}]{personlab}
Papandreou, G.; Zhu, T.; Chen, L.-C.; Gidaris, S.; Tompson, J.; and Murphy, K.
  2018.
\newblock Personlab: Person pose estimation and instance segmentation with a
  bottom-up, part-based, geometric embedding model.
\newblock In \emph{European Conference on Computer Vision}, 269--286.

\bibitem[{Papandreou et~al.(2017)Papandreou, Zhu, Kanazawa, Toshev, Tompson,
  Bregler, and Murphy}]{GooglePose}
Papandreou, G.; Zhu, T.; Kanazawa, N.; Toshev, A.; Tompson, J.; Bregler, C.;
  and Murphy, K. 2017.
\newblock Towards accurate multi-person pose estimation in the wild.
\newblock In \emph{IEEE Conference on Computer Vision and Pattern Recognition},
  4903--4911.

\bibitem[{Paszke et~al.(2019)Paszke, Gross, Massa, Lerer, Bradbury, Chanan,
  Killeen, Lin, Gimelshein, Antiga et~al.}]{Pytorch}
Paszke, A.; Gross, S.; Massa, F.; Lerer, A.; Bradbury, J.; Chanan, G.; Killeen,
  T.; Lin, Z.; Gimelshein, N.; Antiga, L.; et~al. 2019.
\newblock Pytorch: An imperative style, high-performance deep learning library.
\newblock In \emph{Advances in Neural Information Processing Systems},
  8026--8037.

\bibitem[{Pishchulin et~al.(2016)Pishchulin, Insafutdinov, Tang, Andres,
  Andriluka, Gehler, and Schiele}]{DeepCut}
Pishchulin, L.; Insafutdinov, E.; Tang, S.; Andres, B.; Andriluka, M.; Gehler,
  P.~V.; and Schiele, B. 2016.
\newblock Deepcut: Joint subset partition and labeling for multi person pose
  estimation.
\newblock In \emph{IEEE Conference on Computer Vision and Pattern Recognition},
  4929--4937.

\bibitem[{Radosavovic et~al.(2018)Radosavovic, Doll{\'a}r, Girshick, Gkioxari,
  and He}]{dispose1}
Radosavovic, I.; Doll{\'a}r, P.; Girshick, R.; Gkioxari, G.; and He, K. 2018.
\newblock Data distillation: Towards omni-supervised learning.
\newblock In \emph{IEEE Conference on Computer Vision and Pattern Recognition},
  4119--4128.

\bibitem[{Romero et~al.(2014)Romero, Ballas, Kahou, Chassang, Gatta, and
  Bengio}]{fitnets}
Romero, A.; Ballas, N.; Kahou, S.~E.; Chassang, A.; Gatta, C.; and Bengio, Y.
  2014.
\newblock Fitnets: Hints for thin deep nets.
\newblock \emph{arXiv preprint arXiv:1412.6550} .

\bibitem[{Russakovsky et~al.(2015)Russakovsky, Deng, Su, Krause, Satheesh, Ma,
  Huang, Karpathy, Khosla, Bernstein et~al.}]{ImageNet}
Russakovsky, O.; Deng, J.; Su, H.; Krause, J.; Satheesh, S.; Ma, S.; Huang, Z.;
  Karpathy, A.; Khosla, A.; Bernstein, M.; et~al. 2015.
\newblock Imagenet large scale visual recognition challenge.
\newblock \emph{International Journal of Computer Vision} 115(3): 211--252.

\bibitem[{Sun et~al.(2019)Sun, Xiao, Liu, and Wang}]{hrnetv1}
Sun, K.; Xiao, B.; Liu, D.; and Wang, J. 2019.
\newblock Deep high-resolution representation learning for human pose
  estimation.
\newblock In \emph{IEEE Conference on Computer Vision and Pattern Recognition},
  5693--5703.

\bibitem[{Wang et~al.(2019)Wang, Sun, Cheng, Jiang, Deng, Zhao, Liu, Mu, Tan,
  Wang et~al.}]{hrnetpami}
Wang, J.; Sun, K.; Cheng, T.; Jiang, B.; Deng, C.; Zhao, Y.; Liu, D.; Mu, Y.;
  Tan, M.; Wang, X.; et~al. 2019.
\newblock Deep high-resolution representation learning for visual recognition.
\newblock \emph{arXiv preprint arXiv:1908.07919} .

\bibitem[{Wang, Zhao, and Ji(2018)}]{app3}
Wang, K.; Zhao, R.; and Ji, Q. 2018.
\newblock Human computer interaction with head pose, eye gaze and body
  gestures.
\newblock In \emph{IEEE International Conference on Automatic Face \& Gesture
  Recognition}, 789--789.

\bibitem[{Wei et~al.(2016)Wei, Ramakrishna, Kanade, and Sheikh}]{CPM}
Wei, S.-E.; Ramakrishna, V.; Kanade, T.; and Sheikh, Y. 2016.
\newblock Convolutional pose machines.
\newblock In \emph{IEEE Conference on Computer Vision and Pattern Recognition},
  4724--4732.

\bibitem[{Wei et~al.(2018)Wei, Pan, Qin, Ouyang, and Yan}]{disdet3quan}
Wei, Y.; Pan, X.; Qin, H.; Ouyang, W.; and Yan, J. 2018.
\newblock Quantization mimic: Towards very tiny cnn for object detection.
\newblock In \emph{European Conference on Computer Vision}, 267--283.

\bibitem[{Xiao, Wu, and Wei(2018)}]{MSRAPose}
Xiao, B.; Wu, H.; and Wei, Y. 2018.
\newblock Simple baselines for human pose estimation and tracking.
\newblock In \emph{European Conference on Computer Vision}, 466--481.

\bibitem[{Xiu et~al.(2018)Xiu, Li, Wang, Fang, and Lu}]{PoseFlow}
Xiu, Y.; Li, J.; Wang, H.; Fang, Y.; and Lu, C. 2018.
\newblock Pose flow: Efficient online pose tracking.
\newblock In \emph{British Machine Vision Conference}.

\bibitem[{Zhang, Zhu, and Ye(2019)}]{fastpose}
Zhang, F.; Zhu, X.; and Ye, M. 2019.
\newblock Fast human pose estimation.
\newblock In \emph{IEEE Conference on Computer Vision and Pattern Recognition},
  3517--3526.

\bibitem[{Zhang et~al.(2018)Zhang, Xiang, Hospedales, and Lu}]{mutual}
Zhang, Y.; Xiang, T.; Hospedales, T.~M.; and Lu, H. 2018.
\newblock Deep mutual learning.
\newblock In \emph{IEEE Conference on Computer Vision and Pattern Recognition},
  4320--4328.

\bibitem[{Zhang et~al.(2020)Zhang, Peng, Jianlong, Bing, and
  Weiming}]{zhang2020ocean}
Zhang, Z.; Peng, H.; Jianlong, F.; Bing, L.; and Weiming, H. 2020.
\newblock Ocean: Object-aware anchor-free tracking.
\newblock In \emph{European Conference on Computer Vision}.

\bibitem[{Zhou, Koltun, and Kr{\"a}henb{\"u}hl(2020)}]{pointmot}
Zhou, X.; Koltun, V.; and Kr{\"a}henb{\"u}hl, P. 2020.
\newblock Tracking objects as points.
\newblock In \emph{European Conference on Computer Vision}.

\bibitem[{Zhou, Wang, and Kr{\"a}henb{\"u}hl(2019)}]{objects}
Zhou, X.; Wang, D.; and Kr{\"a}henb{\"u}hl, P. 2019.
\newblock Objects as points.
\newblock \emph{arXiv preprint arXiv:1904.07850} .

\bibitem[{Zhu et~al.(2019)Zhu, Hu, Lin, and Dai}]{dcnv2}
Zhu, X.; Hu, H.; Lin, S.; and Dai, J. 2019.
\newblock Deformable convnets v2: More deformable, better results.
\newblock In \emph{IEEE Conference on Computer Vision and Pattern Recognition},
  9308--9316.

\end{thebibliography}

\end{document}